\documentclass{article}
\usepackage{spconf,amsmath,graphicx,cite,booktabs,mathtools,multirow,xcolor}

\newcommand{\red}[1]{\textcolor{red}{#1}}

\usepackage{comment} 
\title{HEAR-YOUR-ACTION: HUMAN ACTION RECOGNITION BY ULTRASOUND ACTIVE SENSING}
\name{Risako Tanigawa and Yasunori Ishii}
\address{Panasonic Holdings Corporation, Yagumonaka-machi, Moriguchi City, Osaka, Japan}

\begin{document}
%
\maketitle
\begin{abstract}
Action recognition is a key technology for many industrial applications. Methods using visual information such as images are very popular.
However, privacy issues prevent widespread usage due to the inclusion of private information, such as visible faces and scene backgrounds, which are not necessary for recognizing user action.
In this paper, we propose a privacy-preserving action recognition by ultrasound active sensing.
As action recognition from ultrasound active sensing in a non-invasive manner is not well investigated, we create a new dataset for action recognition and conduct a comparison of features for classification.
We calculated feature values by focusing on the temporal variation of the amplitude of ultrasound reflected waves and performed classification using a support vector machine and VGG for eight fundamental action classes. 
We confirmed that it is possible to estimate the eight action classes with an accuracy of 97.9\% for the same room and same person for training and evaluation data, and an average accuracy of 62.6\% and a maximum accuracy of 89.5\% for the same room but different persons.
We also report the analyses of accuracies in each condition and limitations.
\end{abstract}
\section{Introduction}
\label{sec:intro}
Action recognition is one of the important technologies that is used for many applications such as robotics \cite{MOLLER2021}, healthcare \cite{SUBASI2020, NWEKE2019, HAQUE2020}, elderly behavior monitoring \cite{DEEP2020, BUZZELLI2020} and suspicious behavior detection \cite{AMRUTHA2020}.
Many of these techniques utilize visual clues, such as RGB videos and images.
Images contain a wealth of visual information about people and scenes.
However, privacy concerns limit the use of scenes that may include identifiable information, such as faces.

To consider the privacy concern, the methods that using radio frequency (RF) signals \cite{WANG2016, LI2019, LI2022}, Wi-Fi signals \cite{GENG2022, BIYUN2020, YAN2020}, and acoustic signals \cite{GAO2020listentolook, IRIE2019SeeingThroughSound, GINOSAR2019, DAS2017GestureUSMicroSoft, MELO2021GestureUS, SHIBATA2023ListeningHumanBehavior} are proposed.
Although RF and Wi-Fi signals can detect fine-grained human postures due to their short wavelengths, the accuracy would degrade due to interference from the electromagnetic waves emitted by electronic devices.
Acoustic signals are also interfered with ambient noise.
However, if the frequency of the target sound is known, it can be restricted by filters.
Furthermore, when using ultrasonic active sensing, the influence of environmental noise is less compared to audible sounds.

The sensing of a person through acoustic signals can be classified into two streams: passive and active sensing.
Passive sensing involves capturing sounds emitted by objects.
Recognition tasks, such as segmentation \cite{IRIE2019SeeingThroughSound} and pose estimation \cite{GAO2020listentolook, GINOSAR2019}, have been performed based on audible sounds, particularly focusing on the voices.
However, voices contain person-identifiable information and can be considered sensitive data from a privacy perspective.
On the other hand, active sensing methods analyze the reflected signals from a person in response to the sounds emitted by the device.
Therefore, these methods allow the acquisition of human movements without using personally identifiable information.
Although the active sensing methods have been applied to gesture recognition \cite{DAS2017GestureUSMicroSoft, MELO2021GestureUS}, segmentation \cite{TANIGAWA2022ECCVW}, and pose estimation \cite{SHIBATA2023ListeningHumanBehavior}, human action recognition by non-invasive ultrasound active sensing has not been well established.
In particular, there are no published action recognition datasets that focus on ultrasonic active sensing.

We propose a new task for human action recognition by ultrasound active sensing.
We build our own datasets.
To construct this dataset, we define 8 basic motion patterns, which include upper-/lower-/whole-body motions and motionless postures.
Then, feature extraction was performed based on time series amplitudes of reflected waves, and action classification was evaluated using a support vector machine and a convolutional neural network.
Our contributions are as follows: (1) We tackle a new task: action recognition by contactless ultrasound sensing; (2) Since there are no previous methods to handle this task, we create a new dataset; and (3) We conduct a comparison of features for classification.

\section{Related Works}
\label{sec:related}

\begin{table*}
    \centering
    \begin{tabular}{cccccc}
        \toprule
        Method & Sensing & Task & Modality & Sound source \\
        \midrule
        Listen to Look \cite{GAO2020listentolook} & Passive & Action recognition & Audio/Visual & Ambient sound \\
        Seeing through Sound \cite{IRIE2019SeeingThroughSound} & Passive & Segmentation & Audio & Subject sound \\
        Conversational gesture recognition\cite{GINOSAR2019} & Passive & 2D hand and arm pose & Audio & Voice \\
        Gesture recognition \cite{DAS2017GestureUSMicroSoft} & Active & Gesture recognition & Ultrasound & Chirp signal \\
        Gesture recognition system\cite{MELO2021GestureUS} & Active & Gesture recognition & Ultrasound & Burst signal \\
        Invisible-to-Visible \cite{TANIGAWA2022ECCVW} & Active & Segmentation & Ultrasound & Burst signal \\
        Listening Human Behavior \cite{SHIBATA2023ListeningHumanBehavior} & Active & 3D Pose & Audible sound & Chirp signal \\
        \bottomrule
    \end{tabular}
    \caption{Comparison of human sensing method using acoustic sensing.}
    \label{table: related works}
\end{table*}

In this section, we present the related work on recognizing action from acoustic sensing.
The related works are summarized in Table \ref{table: related works}.
The acoustic sensing methods are categorized as follows: (1) passive sensing and (2) active sensing.
Passive sensing method is used for action recognition \cite{GAO2020listentolook}, segmentation \cite{IRIE2019SeeingThroughSound}, 2D hand and arm pose \cite{GINOSAR2019}.
These methods use clues of ambient and subject sound including voices.
Voices have been used for speaker recognition tasks therefore passive sensing has privacy concerns due to the usage of voices.
Moreover, passive sensing methods cannot detect human information when the person does not emit any sound such as voices and footsteps.

On the other hand, active sensing allows for the acquisition of human information irrespective of whether a person is emitting sound or not.
The active sensing method is used for gesture recognition \cite{DAS2017GestureUSMicroSoft, MELO2021GestureUS} and 3D pose estimation\cite{SHIBATA2023ListeningHumanBehavior}.
In active sensing, single-frequency burst waves or chirp signals with temporally varying frequencies are employed as acoustic signals for the sound source.
The burst waves are used to measure the distance to objects in \cite{MELO2021GestureUS} and to detect spatial power distribution of reflected waves in \cite{TANIGAWA2022ECCVW}.
The method employing chirp signals enhances the accuracy of Time-of-Flight (ToF) by incorporating signals from multiple frequencies \cite{DAS2017GestureUSMicroSoft}.
Additionally, utilizing the frequency characteristics as features enables the estimation of complex tasks such as 3D pose estimation \cite{SHIBATA2023ListeningHumanBehavior}.

In our approach, we use active sensing for human action recognition to avoid obtaining privacy-related information.
Chirp signals are used for the sound source of our active sensing because they enable us to obtain spatial propagation features in multiple frequencies and are used for other tasks such as gesture recognition and 3D pose estimation.
The frequencies are limited to ultrasound range to avoid capturing voices in the audible range.

\section{Methods}
\label{sec:method}

\begin{figure}[htb]
 \includegraphics[width=\linewidth]{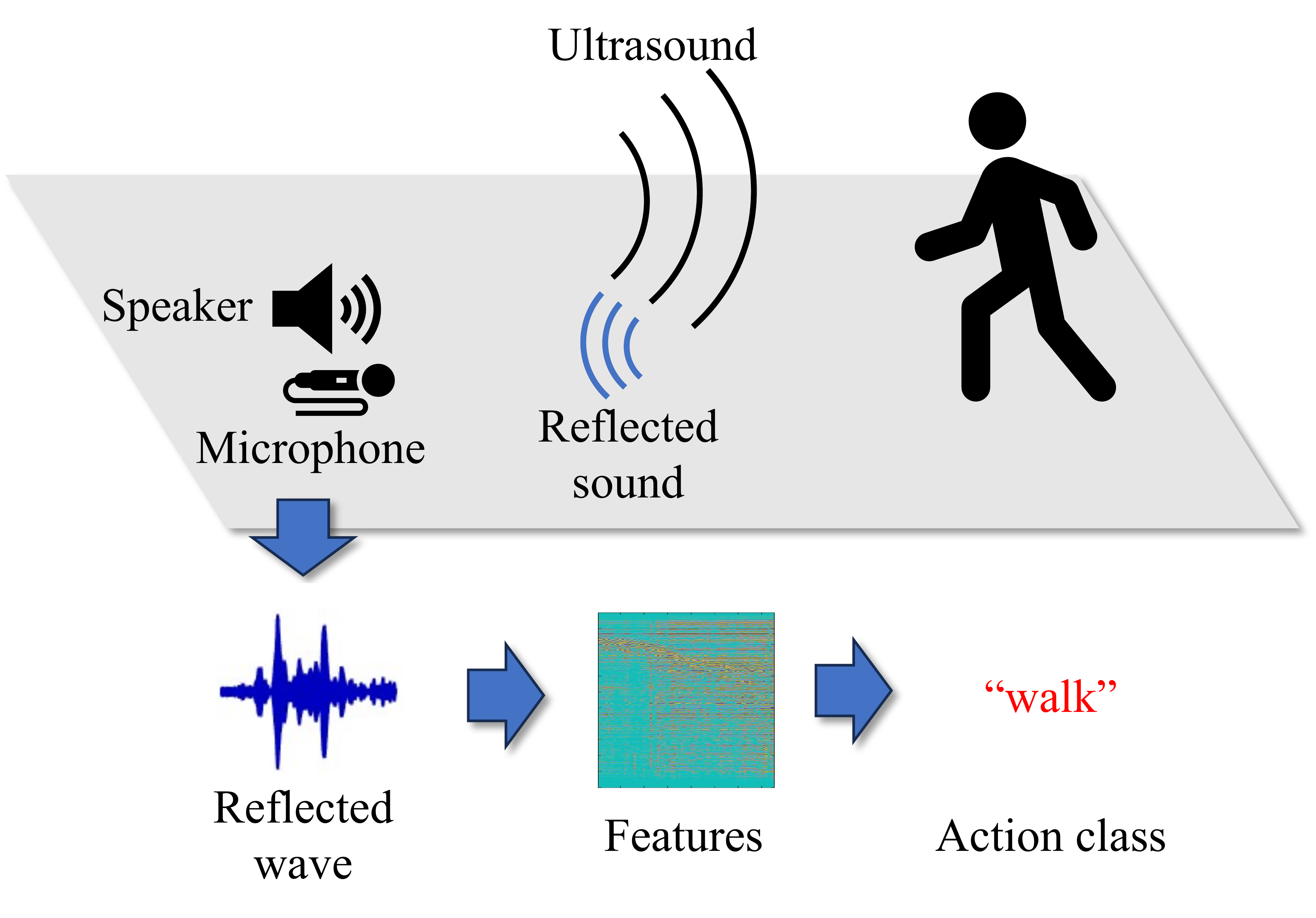}
 \caption{Concept diagram of our approach.}
 \label{fig:concept}
\end{figure}

{\bf Human Action Recognition By Ultrasound Active Sensing:}
We propose a new task that estimates human action from ultrasound signals.
To realize this task, the information we need to know is the changes in the emitted ultrasound from a speaker when it is captured by microphones.
This is the same manner of the room impulse response when using audible sound.
The ultrasound transfer features are changed along with the environment of the space.
If a person moves inside the space, the transfer features of ultrasound would be changed.
Therefore, we established an ultrasound active sensing system for human action recognition and considered feature extraction methods suitable for the action recognition task.
Our concept is described in Fig. \ref{fig:concept}.
We closely set a speaker and microphone opposite to humans and captured reflected signals from the environment and humans.
Analyzed reflected signals to extract features are used for action classifications.

\begin{figure}[htb]
 \includegraphics[width=\linewidth]{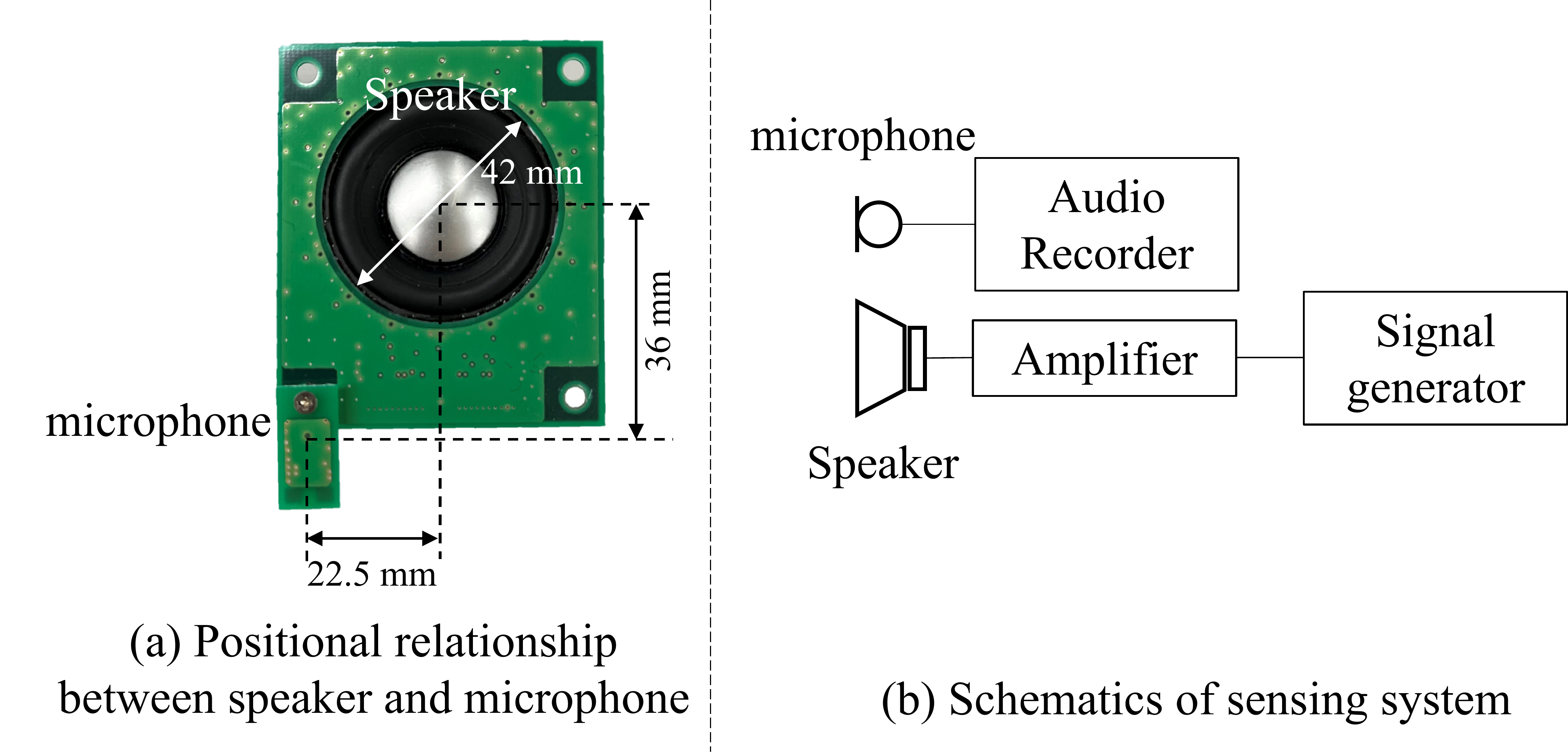}
 \caption{Schematic diagram of the sensing system in our approach.}
 \label{fig:sensingsystem}
\end{figure}

{\bf Active Sensing System:}
The schematic diagram of the active sensing system used in this research is described in Fig. \ref{fig:sensingsystem}.
We used a tweeter to emit ultrasound.
As a receiver, we used two MEMS microphones.
The microphone was placed 38 mm below and 22.5 mm away from the center of the speaker.
The sampling frequency was set to $f_s=96$ kHz enabling to capture of ultrasound range.

{\bf Signal Design:}
We design a linear chirp signal for the active sensing.
The linear chirp signal $x$ is a signal where the frequency increases linearly over time and calculated as
\begin{equation}
	x(t) = \sin \left( 2\pi \left( \frac{\beta}{2} t^2 + f_0 t \right) +\phi_0 \right) ,
\end{equation}
where $t$ is the time, $f_0$ is the lower bound frequency, $\phi_0$ is the initial phase.
$\beta$ is the coefficient determined by the time length of the chirp signals $\tau$ and the lower and upper bound frequencies $f_0$ and $f_1$, respectively:
$
	\beta = (f_1 - f_0)/\tau.
$
In our approach, we set the frequencies to $f_0 = 20$ kHz and $f_1 = 40$ kHz.
To determine the time length of the chirp signal, we considered the minimum distance of the sensing area.
We need to avoid interfering the direct ultrasound and reflected ultrasound when we capture the reflected ultrasound of human.
Therefore, the time length of the chirp signal is limited as $\tau \leq (2 d_\mathrm{min})/c$, where $d_{\mathrm{min}}$ is the minimum distance of the sensing area and $c$ is the speed of ultrasound in air.
We set $d_\mathrm{min}=0.30$ m; therefore, we set $\tau = 1.5$ ms.

To observe the transfer feature changes over time, cyclic chirp signal emission is required.
To do so, we determined the cycle time $T$ of the chirp signals based on the maximum distance of the sensing area $d_\mathrm{max}$: $T \geq (2 d_\mathrm{max})/c$.
We set $d_\mathrm{max}=2.0$ m; therefore, we set $T = 11.8$ ms.

{\bf Feature Extraction:}
While common feature extraction methods have been well-established in audible passive sensing, such as Short-Time Fourier Transform and Mel-Frequency Cepstrum Coefficient, there is limited research in the context of active sensing, and the feature extraction methods are not well-established.
Hence, we considered the following four types of features: (1) time-series reflected waves, (2) time-series envelopes of reflected waves, (3) time-series impulse response of direct and reflected waves, and (4) time-series envelopes of impulse response of direct and reflected waves.

\begin{table*}[t]
    \centering
    \begin{tabular}{c|cccccc|cccc|cccc}
        \toprule
        \multirow{3}{*}{No.} & \multirow{3}{*}{Ra} & \multirow{3}{*}{Rb} & \multirow{3}{*}{Rc(1)} & \multirow{3}{*}{Rc(2)} & \multirow{3}{*}{Rc(3)} & \multirow{3}{*}{Rc(4)} & \multicolumn{8}{c}{Accuracy [\%]} \\
        & & & & & & & \multicolumn{4}{c}{SVM} & \multicolumn{4}{|c}{VGG}\\
        & & & & & & & $F_\mathrm{ref}$ & $F_\mathrm{renv}$ & $F_\mathrm{ir}$ & $F_\mathrm{ienv}$ & $F_\mathrm{ref}$ & $F_\mathrm{renv}$ & $F_\mathrm{ir}$ & $F_\mathrm{ienv}$\\
        \midrule
        1 & T/E & - & - & - & - & - & 87.7 & 91.2 & 90.7 & \textbf{92.6} & \textbf{97.9} & 94.6 & 97.7 & 97.5 \\
        2 & - & T/E & - & - & - & - & 98.1 & \textbf{99.8} & 99.4 & 99.6 & 98.3 & \textbf{99.8} & 97.9 & 99.2 \\
        3 & - & - & E & T & T & T & 80.7 & \textbf{83.6} & 80.7 & 80.1 & 89.3 & \textbf{89.5} & 74.0 & 81.1 \\
        4 & - & - & T & E & T & T & 44.6 & \textbf{62.7} & 42.3 & 59.6 & \textbf{72.1} & 60.0 & 63.3 & 65.6 \\
        5 & - & - & T & T & E & T & 22.2 & \textbf{45.9} & 25.5 & 32.1 & 46.7 & 51.2 & \textbf{56.0} & 42.4 \\
        6 & - & - & T & T & T & E & 1.56 & \textbf{35.5} & 3.91 & 30.5 & 40.2 & \textbf{49.6} & 47.5 & 44.5 \\
        7 & T & T & E & E & E & E & 12.7 & \textbf{14.5} & 10.3 & 11.5 & 22.7 & 22.2 & 21.9 & \textbf{27.6} \\
        \bottomrule
    \end{tabular}
    \caption{Experimental conditions and accuracy results. The character "T" and "E" represents the data used for training and evaluation, respectively.}
    \label{table: acc-result}
\end{table*}

{\bf Time-series reflected waves:}
The first feature value is time-series reflected waves.
Let the received signal of a microphone as $\mathbf{y}$, it includes direct signal and reflected signal.
The direct signal is the wave that directly approaches from the speaker, and the reflected signal is the wave propagated to the space and reaches the microphones.
Since the reflected waves propagate through the space occupied by a person, we extract each reflected wave from a single chirp signal and concatenate them as follows: 
$
    F_\mathrm{ref} = [\mathbf{y}_1, \mathbf{y}_2, \ldots, \mathbf{y}_N]
$, where $F_\mathrm{ref}$ is the time-series reflected waves and $\mathbf{y}_i \, (i = 1, 2, \ldots, N)$ is the $i$-th reflected wave: 
$
    \mathbf{y}_i = \mathbf{y}[N_{\mathrm{dir},i}+N_\mathrm{min}, \ldots, N_{\mathrm{dir},i}+N_\mathrm{max}]
$, where $N_{\mathrm{dir},i}$ is the index of $i$-th direct wave, $N_\mathrm{min} = 2 f_s d_\mathrm{min} / c$ is the index corresponding to the minimum distance of the sensing area $d_\mathrm{min}$, and $N_\mathrm{max} = 2 f_s d_\mathrm{max} /c$ is the index corresponding to the maximum distance of the sensing area.

{\bf Time-series envelopes of reflected waves:}
The second feature value is the time-series envelopes of reflected waves.
Since the phase of reflected waves greatly depends on the shape of the reflecting object, it is considered as environment-dependent information.
To focus on the amplitude of reflected waves, the envelopes of the amplitude of each reflected wave are extracted and concatenated as the time-series envelopes of the reflected waves $
    F_\mathrm{renv}$: $F_\mathrm{renv} = [\mathbf{\hat{y}}_1, \mathbf{\hat{y}}_2, \ldots, \mathbf{\hat{y}}_N]
$, where $\mathbf{\hat{y}}_i \, (i = 1, 2, \ldots, N)$ is the $i$-th envelope of reflected wave.

{\bf Time-series impulse response of direct and reflected waves:}
The third feature value is the time-series impulse response of direct and reflected waves.
Both the direct wave and reflected wave are emitted from the same speaker and received by the same microphone.
Therefore, by calculating the impulse response of the direct wave and reflected wave, the signals can be obtained that eliminate the characteristics of the speaker and microphone.
When $Y_\mathrm{dir}$ and $Y_\mathrm{ref}$ represent the frequency-domain signal of the direct and reflected wave, respectively, the transfer function $H_\mathrm{rd}$ of the direct wave and reflected wave can be expressed as 
$
    H_\mathrm{rd}(\omega) = Y_\mathrm{ref}(\omega)/Y_\mathrm{dir}(\omega)
$, where $\omega$ is the angular frequency.
The feature value of the time-series impulse response of direct and reflected waves is formed by concatenating the $i$-th time-domain signal $h_{\mathrm{rd},i}$ of the transfer function $
    H_{\mathrm{rd},i}$: $F_\mathrm{ir} = [\mathbf{h}_{\mathrm{rd},1}, \mathbf{h}_{\mathrm{rd},2}, \ldots, \mathbf{h}_{\mathrm{rd}, N}]
$.

{\bf Time-series envelopes of impulse response of direct and reflected waves:}
The last feature value is the time-series envelopes of impulse response of direct and reflected waves.
The value is formed as
$
    F_\mathrm{ienv} = [\mathbf{\hat{h}}_{\mathrm{rd},1}, \mathbf{\hat{h}}_{\mathrm{rd},2}, \ldots, \mathbf{\hat{h}}_{\mathrm{rd}, N}],
$
where $\mathbf{\hat{h}}_{\mathrm{rd}, i}$ is the $i$-th envelope of the impulse response of direct and reflected waves.

\section{Experimental Result}
\label{sec:exp}

\begin{figure}[htb]
 \includegraphics[width=\linewidth]{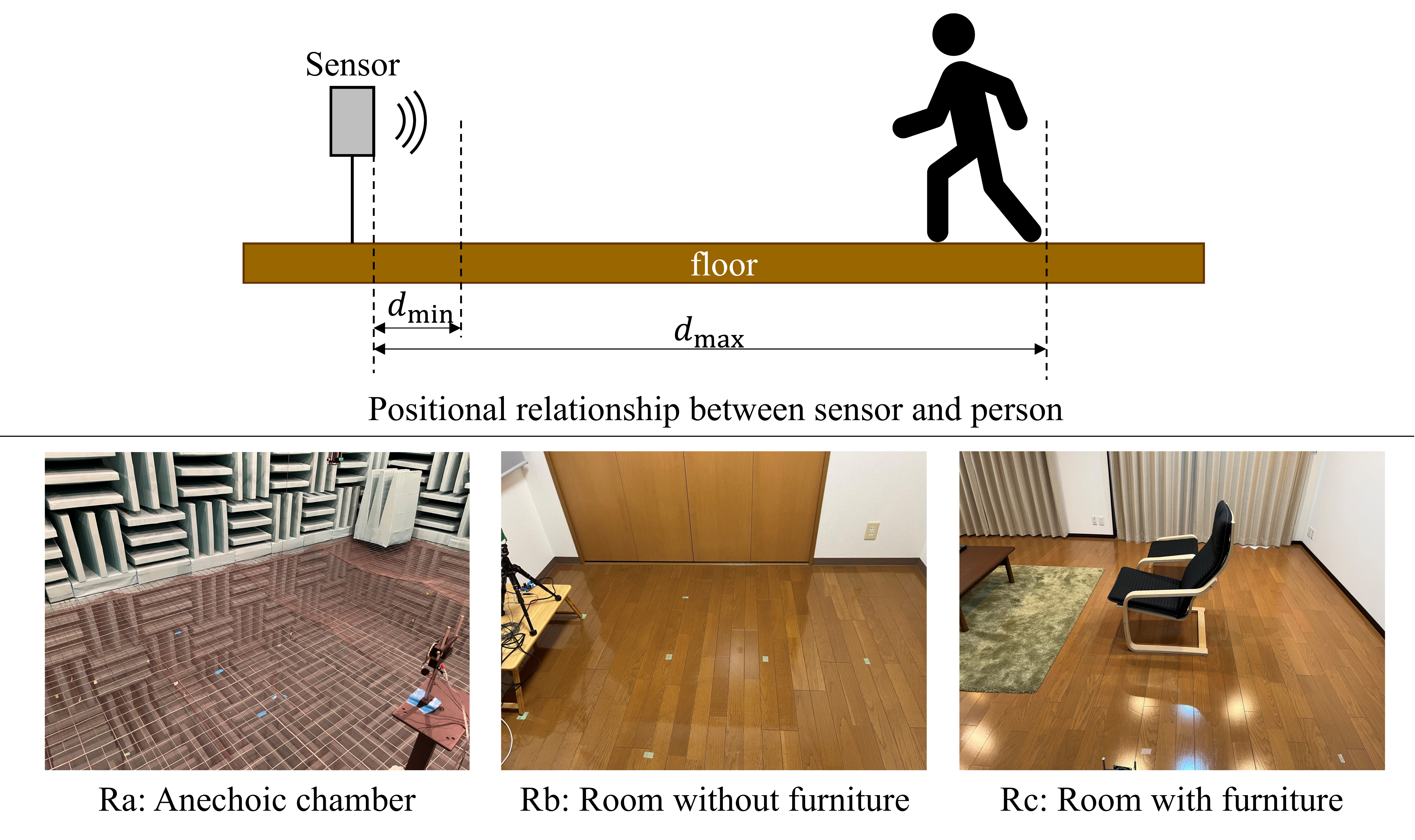}
 \caption{Schematic diagram of data acquisition condition and pictures of room.}
 \label{fig:exp_room}
\end{figure}

\subsection{Datasets}
Since there \red{have been} no existing ultrasound action recognition datasets available, we proceeded to create our own dataset to evaluate the possibility of action recognition from ultrasound.
The experimental setup is described in Fig. \ref{fig:exp_room}.
We used the tweeter as a sound source and the MEMS microphone in this experiment.
We recorded the data in three different spaces: (Ra) anechoic chamber, (Rb) a room without furniture, (Rc) a room with furniture.
In the anechoic chamber and a room without furniture, we recorded data with one subject.
In the room with furniture, we recorded single-person data with four subjects.
We set eight fundamental action classes: {\tt hand-waving}, {\tt throwing}, {\tt kicking}, {\tt picking-up}, {\tt walking}, {\tt lying-down}, {\tt sitting}, and {\tt standing}.

\subsection{Settings}
To classify action classes from ultrasound features, we \red{use} two models: linear support vector machine (SVM) \cite{CORTES1995SVM} and VGG\cite{SIMONYAN2015VGG}.
We \red{use} a VGG consisting of $19$ layers without pretraining.
We set the learning rate to $0.1$ and \red{use} a batch size of $12$.
The optimizser is the Stochastic Gradient Descent (SGD) with a momentum of $0.9$ and a weight decay of $0.0005$.
\red{The input tensor size is hoge $\times$ hoge $\times$ 1.}
\red{we also use data augmentation, which is rotation within $\pm$5 degree and horizontal flip.}
\red{we use the accuracy score for all evaluations.}

\subsection{\red{Results}}
\label{sec:res}
\red{
Table \ref{table: acc-result} shows the evaluation result.
}
Conditions No. 1 and 2 represent the simplest arrangement where we use \red{the same room and same person, however, the data is split between training and evaluation.}
The accuracy was reached to 99.8\% in the best condition.
\red{
From this result, we confirmed that SVM and VGG have approximately the same accuracy and are highly accurate.
Therefore, under ideal conditions, actions can be classified with the proposed waveform.
}
Conditions from No. 3 to 6 evaluated the accuracy for unknown subjects.
We split the data of Rc into four sets per subject and evaluated by k-means cross-validation, where $k=4$.
By referring to No. 3 to 6 in Table \ref{table: acc-result}, there is variation in accuracy depending on the set used for training and evaluation.
\red{These results show that the waveforms vary between people, depending on the speed and motion.
When the intra-class variance of each action in the training data is large, VGG can learn discriminative boundaries that are more useful for classification than linear SVM.
Thus, compared to the accuracies of SVM, the accuracies of VGG tends to be higher.}

Condition No. 7 evaluated the accuracy for the unknown room.
The accuracy sharply dropped and reached a mere 27.6\%, even at its highest score.
This is because the reflected waves include reflections not only from humans but also from objects other than humans.
Since these reflections vary depending on the environment, the robustness against data from environments is considered to be low.
\red{Our experiments revealed that the proposed action recognition method needs to be more robust to unknown environments.
In order to improve the issue, it is effective to standardize the data using data when no one is present, or to remove the stationary component by taking the difference in reflected waves for each chirp signal.
Alternatively, by collecting large amounts of data and performing deep learning-based learning, the performance can be expected to improve by learning a recognizer that is robust to differences between people and the environment.}


\section{\red{Conclusions}}
\label{sec:conclusion}


\vfill
\pagebreak

\bibliographystyle{IEEEbib}
\bibliography{strings,refs}

\end{document}